\pdfoutput=1

\documentclass[11pt]{article}

\usepackage{acl} 
\usepackage{times}
\usepackage{latexsym}
\usepackage{tabularx}
\usepackage{multibib}
\usepackage{multirow}
\usepackage{xcolor,colortbl}
\usepackage{mathtools}  
\usepackage{makecell} 
\usepackage{booktabs}
\usepackage{adjustbox}
\usepackage{footnote}
\usepackage{linguex}
\usepackage{cleveref}

\crefformat{section}{\S#2#1#3} 
\crefformat{subsection}{\S#2#1#3}
\crefformat{subsubsection}{\S#2#1#3}

\usepackage[T1]{fontenc}

\usepackage[utf8]{inputenc}

\usepackage{microtype}

\title{Summarizing Patients' Problems from Hospital Progress Notes Using Pre-trained Sequence-to-Sequence Models}

 \author{Yanjun Gao* \\ ICU Data Science Lab \\  School of Medicine and Public Health \\ University of Wisconsin-Madison \\ \texttt{ \small ygao@medicine.wisc.edu}  
         \And
         Dmitriy Dligach \\ Loyola University Chicago \\ \texttt{ \small ddligach@luc.edu}   \AND 
         Timothy Miller, Dongfang Xu \\ Boston Children's Hospital \\ and Harvard Medical School \\ \texttt{ \small Timothy.Miller, Dongfang.Xu} \\ 
         \texttt{ \small @childrens.harvard.edu} \And 
         Matthew M. Churpek, Majid Afshar \\ ICU Data Science Lab \\  School of Medicine and Public Health \\ University of Wisconsin-Madison \\ 
         \texttt{ \small mchurpek, mafshar@medicine.wisc.edu }  
         }

\begin{document}
\maketitle

\begin{abstract}
Automatically summarizing patients' main problems from daily progress notes using natural language processing methods helps to battle against information and cognitive overload in hospital settings and potentially assists providers with computerized diagnostic decision support. Problem list summarization requires a model to understand, abstract, and generate clinical documentation. In this work, we propose a new NLP task that aims to generate a list of problems in a patient's daily care plan using input from the provider's progress notes during hospitalization. We investigate the performance of T5 and BART, two state-of-the-art seq2seq transformer architectures, in solving this problem. We provide a corpus built on top of progress notes from publicly available electronic health record progress notes in the Medical Information Mart for Intensive Care (MIMIC)-III. T5 and BART are trained on general domain text, and we experiment with a data augmentation method and a domain adaptation pre-training method to increase exposure to medical vocabulary and knowledge. Evaluation methods include ROUGE, BERTScore, cosine similarity on sentence embedding, and F-score on medical concepts. Results show that T5 with domain adaptive pre-training achieves significant performance gains compared to a rule-based system and general domain pre-trained language models, indicating a promising direction for tackling the problem summarization task.      
\end{abstract}

\section{Introduction}
The progress note is a common note type in the electronic health record (EHR) that also contains the necessary details for medical billing; therefore, every hospital day will contain at least one progress note for a patient. Healthcare providers write them to document a patient's daily progress and care plan~\cite{brown2014physicians}. The progress note contains both subjective and objective information gathered by the care team, and it is updated daily and serves as the most viewed clinical document by providers. The complexity of the progress note increases as the patient's illness worsens with progress notes collected in the intensive care unit (ICU) representing the sickest patients in the hospital. In the ICU, information and cognitive overload occur frequently, with more opportunities for missed diagnoses and medical errors~\cite{furlow2020information, hultman2019challenges}. Automatically generating a set of diagnoses/problems in a progress note may assist providers in overcoming cognitive biases and heuristics and apply evidence-based medicine via information synthesis to accurately understand a patient's condition. These processes may ultimately reduce the effort in document review and augment care during a time-sensitive hospital event~\cite{devarakonda2017automated}.  

\begin{figure}
    \centering
    \includegraphics[width=\columnwidth]{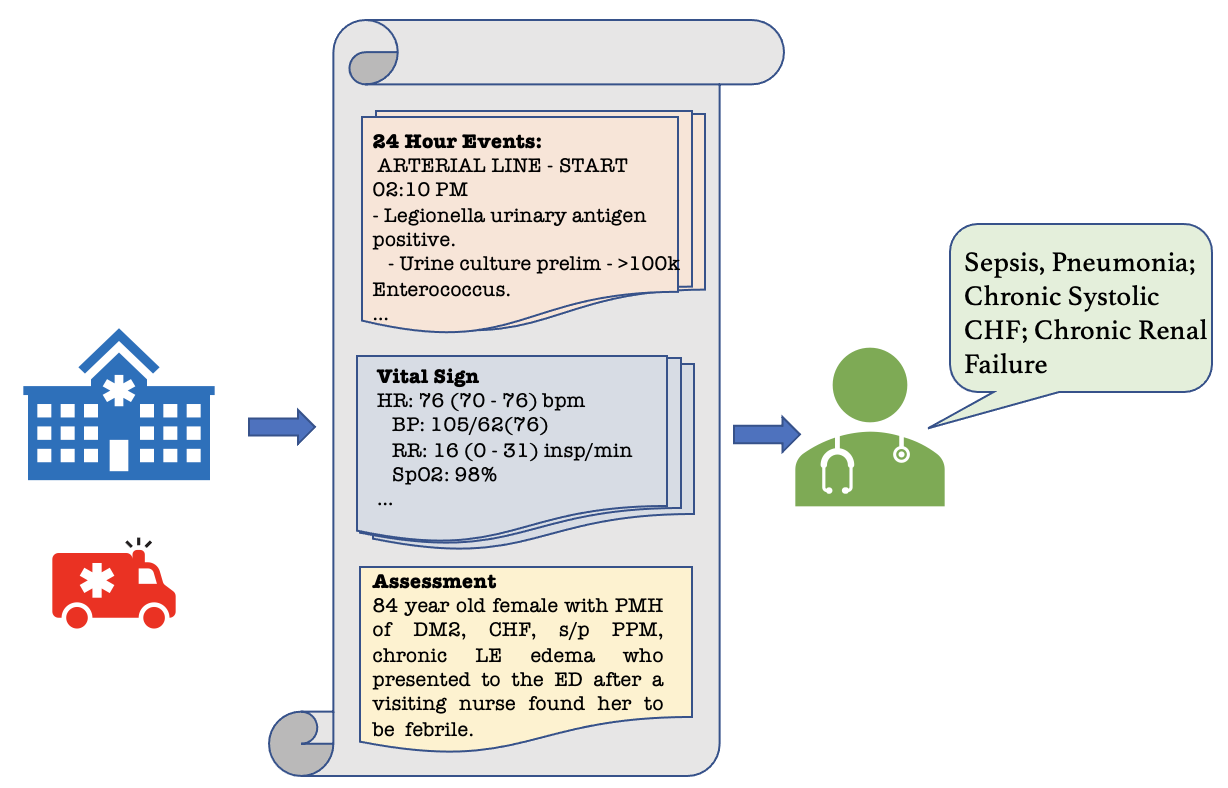}
    \vspace{-.2in}
    \caption{\small When a sick patient arrives to the hospital, diagnostic evaluations are performed to assess the patient's condition and deduce the problems causing the illness.  }
    \label{fig:hospital}
\end{figure}

Clinical note summarization using natural language processing (NLP) has demonstrated promise in previous work. \citeauthor{hirsch2015harvest} (\citeyear{hirsch2015harvest}) introduced HARVEST, an EHR summarizer that is currently deployed at point-of-care in a New York hospital. The NLP components of HARVEST include a Markov chain named-entity tagger that identifies diseases explicitly mentioned in clinical notes and a TF-IDF scorer that weighs the importance of the mentions~\cite{lipsky2011clinnote,hirsch2015harvest}. With the advances of neural methods, recent work has focused on radiology report summarization~\cite{zhang2018learning, macavaney2019ontology, gharebagh2020attend} with pointer generator network~\cite{see2017get}, and doctor-patient conversation summarization~\cite{yim2021towards,zhang2021leveraging} with transformer architectures~\cite{vaswani2017attention,raffel2020exploring}. Few investigations apply transformers to \textit{Problem Summarization}  progress notes to identify and generate the top diagnoses during a patient's hospitalization.     

Problem summarization requires complex cognitive processes to arrive at an accurate diagnosis. When a patient is admitted to the hospital, medical evaluations and diagnostics are initially performed to understand a patient's condition. The review is accompanied by documentation in the progress notes to include pertinent details about the patient's symptoms, medications, physical exam findings, radiology findings, laboratory results, etc. These data are organized in the progress note and used with the physician's medical knowledge to arrive at an assessment of the current problems followed by a treatment plan. The system of nonanalytic and analytic reasoning strategies represent \textbf{clinical diagnostic reasoning}, a process involving clinical evidence acquisition with integration and abstraction over medical knowledge to synthesize a conclusion in the form of a diagnosis\cite{barrows1980problem, bowen2006educational}. 
We hypothesize that to summarize a patient's problems and ultimately develop computerized diagnostic decision support systems, the ability of clinical diagnostic reasoning is the key for NLP systems, a gap in existing NLP literature. In this work, we propose a new summarization task designed to meet the real-world need in the hospital setting as the first step to developing NLP models for clinical diagnostic reasoning. The task is built on a new annotation subset of MIMIC-III~\cite{johnson2016mimic}, a large and publicly available EHR. We formulate the task as a problem list summarization, as we see the task as the first step in a bigger vision of generating entire notes or sections of notes. Ultimately, the task is designed with our clinical informatics partners to move toward a future real-world application, where a system generating relevant diagnoses can assist healthcare providers and overcome the cognitive burden and information overload.  
Our contributions include: 
\begin{itemize}
    \item The first knowledge-intensive summarization task towards building NLP systems for computerized diagnostic decision support (sec \cref{sec:task}), with an annotated set of clinical notes that are publicly available (sec \cref{sec:data});
    \item An evaluation of two transformer models for this task, T5~\cite{raffel2020exploring} and BART~\cite{lewis2020bart}, to examine progress in using the state-of-the-art models over a rule-based medical concept extractor (sec \cref{sec:setting});  
    \item Domain adaptive pre-training to establish benchmark performance for this task across multiple evaluation metrics (sec \cref{sec:results}), with discussion of key challenges and future directions (\cref{sec:discussion}).  
\end{itemize}

\section{Related Work}
In this section, we provide a brief overview of recently published papers on clinical summarization that use neural methods.  

\paragraph{Task setup} The stream of recent work on clinical summarization may be divided into two groups: extractive summarization and abstractive summarization. The data corpora are heterogeneous, with multiple note types represented. For extractive sumarization, \citeauthor{liang-etal-2019-novel-system} (\citeyear{liang-etal-2019-novel-system}) proposes a summarization task that extracts sentences from progress notes. \citeauthor{adams2021s}(\citeyear{adams2021s}) introduces a clinical note summarization task to generate a discharge summary generated from prior notes during hospitalization. More efforts have been made toward abstractive summarization. Several work focus on summarizing radiology reports into an impression, a short piece of text stating the findings from the source image~\cite{zhang2018learning,macavaney2019ontology,gharebagh2020attend}. Another task is doctor-patient conversation summarization where the output is a summary describing the patient's visit:~\cite{yim2021towards, manas2021knowledge,zhang2021leveraging}; or generating clinical notes using both extractive and abstractive summarization: \cite{krishna2021generating}. Our work is similar to \cite{liang-etal-2019-novel-system} in the emphasis on summarizing problems from progress notes. Yet, \citeauthor{liang-etal-2019-novel-system} (\citeyear{liang-etal-2019-novel-system}) uses a disease-specific dataset (hypotension and diabetes), and formulates the problem as extractive summarization. Our annotations span a broad range of diagnoses across multiple disciplines (surgery, medicine, neurology, cardiology, trauma, etc.) and investigate  extractive and abstractive approaches in the task. 

\paragraph{Evaluation} Prior work has relied on ROUGE~\cite{lin2004rouge} as the primary evaluation metric for summarization. Most papers also report human evaluation with aspects of clinical relevancy, factual accuracy and readability~\cite{macavaney2019ontology,gharebagh2020attend,krishna2021generating,yim2021towards,zhang2021leveraging}. Few have evaluated using a concept F-score, measuring if the predicted summaries contain accurate medical concepts~\cite{liang-etal-2019-novel-system, zhang2021leveraging}. Our work follows prior work and uses ROUGE, concept F-score, and human evaluation to assess the quality of generated summaries. We also evaluate content quality based on semantic representation using BERTScore~\cite{bert-score} and cosine similarity for sentence embedding.

\section{Task Description} \label{sec:task}


\begin{figure}[t!]
\centering
\begin{adjustbox}{width=1\columnwidth} 
\footnotesize 
    \centering
  \begin{tabular}{p{\columnwidth}}  \hline 
      \textbf{Input Assessment and Subjective Sections} \\  
        (\textit{Assessment}) Pt is a 78 y.o female with h.o COPD, HTN, recent MVA with R.ankle/foot fx who presents with hypoxia and LLL infiltrate. \\ 
        
        \textit{Chief Complaint}: Pt does not feel better than at admission, still very fatigued and weak. SOB unchanged. No chest pain. No other complaints. \\
        \textit{Allergies}: No Known Drug Allergies \\
        \textit{Review of systems} is unchanged from admission except as noted below Review of systems: \\ \hline 
  \end{tabular}
  \end{adjustbox}
   \vspace{-.14in}
    \caption{\small An input example of assessment and subjective sections available in the notes: Chief Complaint, Allergies, Review of systems.}
    \label{fig:a_subj}
\end{figure}

Many clinical NLP applications aim to improve physicians' efficiency and decision-making by automatically highlighting essential information from the large body of textual data in the EHR. The goal of Problem Summarization is to identify and generate the problems and diagnoses for the patient's ICU stay. The Problem Summarization task could be developed using a multi-document approach with all notes captured during a hospital encounter. A patient encounter may generate multiple clinical notes (e.g. admission note, transfer note, daily progress notes, etc.), involving different modalities of data such as structured EHR data and radiology images. However, we are particularly interested in facilitating NLP model development for clinical diagnostic reasoning. We define the task as single-document summarization and focus only on a cross-sectional point in time with a single progress note. Our work will show that summarizing a patient's problems over a single progress note is a challenging task and a necessary foundation that requires clinical text understanding and reasoning over sequences of medical concepts.

The progress note is organized in the ubiquitous SOAP format with four components: \textit{S}ubjective, \textit{O}bjective, \textit{A}ssessment, and \textit{P}lan, a documentation method designed to present patient's problems in a highly structured way and developed by Larry Weed, MD~\cite{weed1964medical}. Each component has multiple sections gathering patients' information, helping the healthcare providers quickly recognize medical events and active problems. \textsc{Subjective} sections are written in natural language and record information about health concerns expressed by patients (e.g. \textsc{Chief Complaints}), and past medical events and history (e.g. \textsc{Allergies, Family History}). \textsc{Objective} sections are primarily structured data, including vital signs, lab tests, medications. \textsc{Assessment} is a brief description of passive and active diagnoses. It states why the patient is admitted to the hospital and the active problem for the day, usually accompanied by the patient's comorbidities. \textsc{Plan} section includes multiple subsections, each listing a medical problem and treatment plan. The progress note is time-sensitive EHR data because it is documented daily. As a patient’s condition changes and the length of stay increases, the progress note may also increase in length. Another reason for the increasing size is from copy-and-paste behaviour, also known as ``note bloat" adding redundant information or noise and hindering the efficiency in data synthesis, which increases the risk for medical error~\cite{rule2021length,tsou2017safe,shoolin2013association}. This reiterates our motivation to develop an NLP system that automatically generates problems and diagnoses to assist providers in clinical workflow and improve diagnostic accuracy.  

Our task took Subjective and Assessment sections in progress notes as input and omitted the Objective sections. Both the Subjective sections and the Assessment section contained information about the reason for admission; therefore, they became the source text (see Figure~\ref{fig:a_subj} for an example). The reference summary is a list of problems mentioned in each Plan subsection relevant to the reasons for hospitalization. We will explain the annotation process in the next section. \footnote{Training script is available at: \url{https://git.doit.wisc.edu/smph/dom/UW-ICU-Data-Science-Lab/drbench}.}

\begin{figure*}[t]
    \centering
    \includegraphics[scale=0.76]{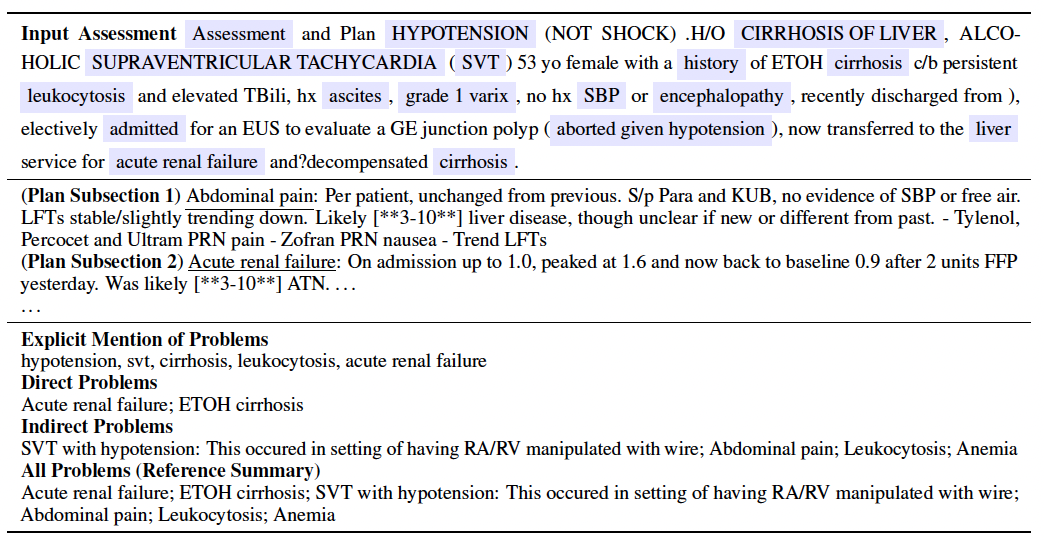}
    \vspace{-.15in}
    \caption{\small \textbf{Top}: An example assessment input with all the concepts (highlighted in \colorbox{blue!10}{color box}) identified through QuickUMLS, a state-of-the-art off-the-shelf medical concept extractor. \textbf{Middle}: Two example plan subsections with the \underline{the annotated problems}, with relation labels omitted. \textbf{Bottom}: The reference summary (All Problems) consists of problems annotated as the main reasons for hospitalization (Direct Problems) and secondary concerns (Indirect Problems); explicit mention of the problems is detected by overlapping the concepts identified through UMLS in input and reference summary. } 
    \label{fig:task}
\end{figure*}

\section{Data} \label{sec:data}
All progress notes were sourced from MIMIC-III, a publicly available dataset of de-identified EHR data from approximately 60,000 hospital ICU admissions at Beth Isreal Deaconess Medical Center in Boston, Massachusetts. We randomly sampled a subset of 768 progress notes and annotated the text spans for the SOAP components. The goal of the annotation was to obtain lists of problems from the Plan subsections. For each Plan subsection, the annotators marked the text span for the Problem, separating the diagnosis/problems from the treatment or action plans. The annotators subsequently determined if the problem was a primary diagnosis (\textit{Direct}), or a past medical problem or consequence from the primary diagnosis (\textit{Indirect}). Two more labels were available for annotating the Plan subsection: \textit{Neither} if the problem or diagnosis was not mentioned in the progress note;  \textit{Not Relevant} if the Plan subsection contained non-diagnostic comments such as describing nutrition, prophylaxis, or disposition. Finally, we concatenated the Direct and Indirect problems using semi-colons and used them as \textit{reference summary}. Two medical school students were trained as annotators under the supervision of two board-certified critical care ICU physicians. On the four labels, they achieved a Cohen’s Kappa of 0.74 on 10 randomly sampled notes, considered as good quality given the complexity of the task. More details may be found in the Appendix~\ref{appendix:annotation}. \footnote{Annotation is available through PhysioNet.}

Figure~\ref{fig:task} illustrates the task setup. The Direct and Indirect problems 
were labeled from each Plan subsection using information presented in the input Assessment (entire progress note was also available to the annotators for more information), forming the reference summary (All Problems in the bottom). 
A total of 1404 and 1599 text spans were labeled as Direct and Indirect Problems, respectively. The majority of the Direct problems were found in the input Assessment but many of the Indirect problems were not explicitly mentioned in the input Assessment and may be found in other parts of the progress note (\textit{abdominal pain} finding in Subjective or \textit{pneumothorax} finding in chest imaging result of Objective). We also performed medical concept mapping through UMLS (see \cref{sec:setting}) on the input Assessment and kept the overlap with the reference summaries and categorized them \textit{Explicit Mention of Problems} as an automated labeling approach and baseline. Therefore, the problems represent extractive and abstractive medical concepts. We presented the results across these subgroups assuming the complexity increases as we move from Explicit to Direct to Indirect problems.

\section{Experiment Setting} \label{sec:setting}
The Unified Medical Language System (UMLS) from the National Library of Medicine is the largest resource containing biomedical concepts and their relationships~\cite{bodenreider2004unified}. We applied the concept extractor from QuickUMLS~\cite{soldainiquickumls}, a fast and lightweight Python package, to identify all the medical concepts in the text as the baseline system. Two state-of-the-art seq2seq transformers were selected to compare with the rule-based method: T5~\cite{raffel2020exploring} and BART~\cite{lewis2020bart}. The transformer models are known as data hungry and pre-trained on domain general text, yet our training data was limited in size but full of medical terms. To help the model learn the medical vocabulary and knowledge, we used data augmentation to generate more training samples for our experiments (\cref{subsec:data_aug} and \cref{subsec:dapt_method}).   

\subsection{Data augmentation} \label{subsec:data_aug}

\begin{figure}[!t]
    \centering
    \includegraphics[width=\columnwidth]{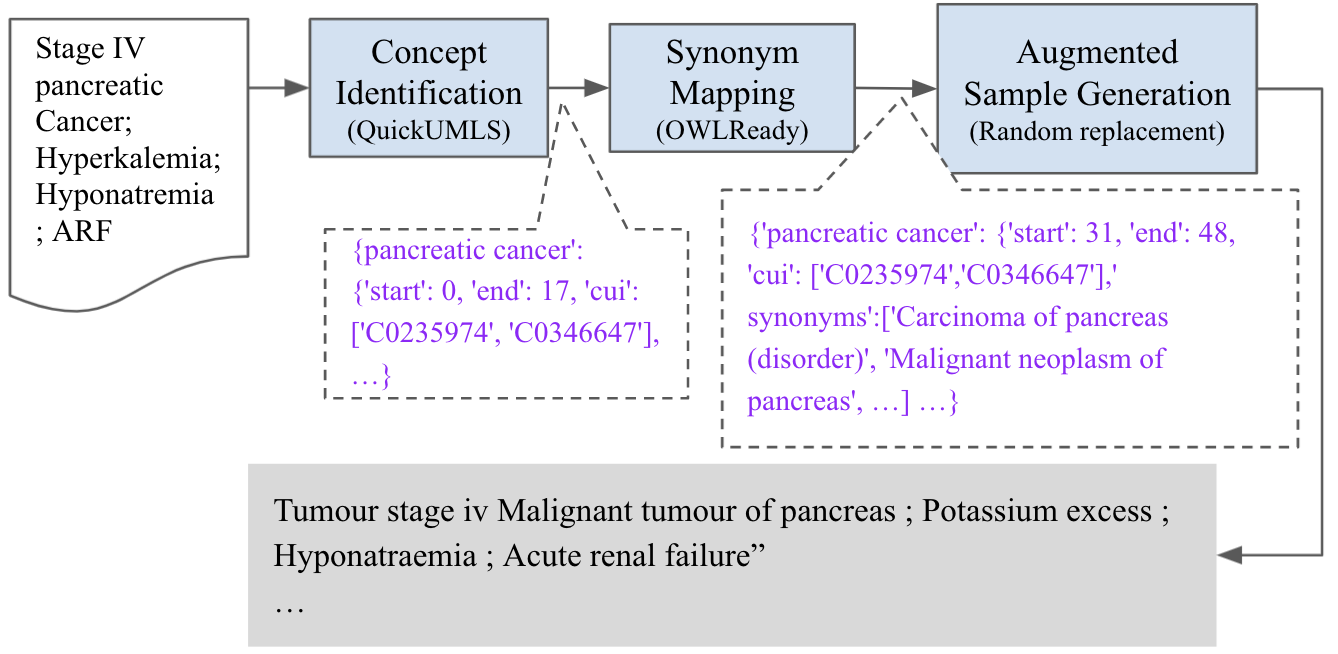}
     \vspace{-.28in}
    \caption{\small Workflow of the data augmentation method with an input reference summary and output augmented sample}
    
    \label{fig:augment_flow}
\end{figure}

Figure~\ref{fig:augment_flow} presents a workflow of the data augmentation method across the following three steps: (1) concept identification; (2) synonym mapping; and (3) augmented sample generation. Given an input text, the step of concept identification extracted ngram terms that were matched concepts with UMLS entities, from QuickUMLS. This step was done through a text matcher algorithm using a cosine similarity threshold, setting as Jaccard score with cosine similarity as 1 in our use case. 
The results returned Concept Unique Identifiers (CUI), a symbolic ID for the medical concept from UMLS. 
An example output of this step is illustrated in Figure~\ref{fig:augment_flow}: a dictionary of the matched ngrams, e.g. \textit{``pancreatic cancer''}, with start and end character positions and CUIs, e.g. [C0235974]. 
The mapping module in step 2 found synonyms through CUIs. Here, we used OWLReady~\cite{lamy2017owlready} that automatically constructed an UMLS ontology graph, linking the concepts with relations and enabling a quick synonym lookup given a CUI. The synonyms were then 
passed to the last module for augmented sample generation. The last module randomly chose the synonyms and replaced concepts. An input text may contain $n$ concepts, with each concept having $r$ maximum number of synonyms, the number of combinations of synonyms $r^n$ grows exponentially as $n$ increases. Considering the efficiency, we limited the number of combinations generated by concept replacement to 1000. 
We ran the pipeline on both reference summary and input assessment, and obtained approximately 132,000 pairs of samples for additional training data. We conducted quality measurement on the augmented samples and report the results in the Appendix~\ref{appendix:quality}.  


\begin{table}[!t]
    \small 
    \centering
    \begin{tabular}{clll} \toprule 
       Set  &  Fine-tuning & Data Augmt. & DAPT  \\ 
       \midrule 
       \#Notes & 700 & 132k & 293k \\  
       Input Lens$_{(\sigma)}$ & 43.33$_{24.75}$ & 212.74$_{78.39}$ & 46.19$_{70.95}$   \\  
        \bottomrule 
    \end{tabular}
    \vspace{-.1in}
    \caption{\small Size and average input length (and standard deviation $\sigma$) of training set for different experiment settings: the original annotated set for fine-tuning, the data generated from data augmentation method, and DAPT. }
    \label{tab:train_size}
\end{table}

\subsection{Domain adaptive pretraining with random concept masking} \label{subsec:dapt_method}
The summarization task requires clinical text understanding and medical knowledge, exposing challenges for models pre-trained from the general domain. Previous work proposed strategies of continuously training the pre-trained language model on domain-specific tasks to enable domain knowledge learning.~\cite{gupta2021effect,pruksachatkun2020intermediate,gururangan2020don}. We followed a similar approach to investigate the effect of domain adaptive pre-training (DAPT) on our summarization task. Specifically, we continuously trained T5 on \textsc{Assessment and Plan} sections from all progress notes in MIMIC, excluding the set of notes for the test set. The result set had 293,000 notes, with the top 3 most frequent note types as Nursing Progress Notes (181k), Physician Resident Progress Notes (61k), and Intensivist Notes (25k). 

T5 was trained by random token masking: given a text string, it randomly replaced the token spans with a special tag \textit{``<extra\_id\_>"} and learned to generate the masked tokens. However, not all words were equally important in our task and we wanted the model to learn clinical semantic types such as symptoms and diseases. Previous work proposed masking on biomedical entities and time expression, achieving performance gains when compared to BERT without entity masking~\cite{lin2021entitybert, pergola2021boosting}.  
Inspired by these work, we adapted a concept masking policy where we randomly masked the concepts identified through UMLS. We set the mask token ratio to 15\%. For example, the highlighted text in Figure~\ref{fig:task} was randomly replaced with the special tag. The statistics of the training set are shown in Table~\ref{tab:train_size}. 

\begin{table*}[hbt!]
\begin{adjustbox}{width=1\textwidth} 

\small 
    \centering
    \begin{tabular}{cll>{\columncolor[gray]{0.9}}ll>{\columncolor[gray]{0.9}}l|
    l>{\columncolor[gray]{0.9}}ll>{\columncolor[gray]{0.9}}l|
    l>{\columncolor[gray]{0.9}}ll>{\columncolor[gray]{0.9}}l|
    l>{\columncolor[gray]{0.9}}ll>{\columncolor[gray]{0.9}}l} \toprule 
         \multirow{2}{*}{\textbf{Model}} & 
         \multirow{2}{*}{\textbf{Setting}} & 
        \multicolumn{4}{c|}{\textbf{Explicit Mentions}} &
         \multicolumn{4}{c|}{\textbf{Direct Problems}} &
          \multicolumn{4}{c|}{\textbf{Indirect Problems}} &
         \multicolumn{4}{c}{\textbf{All Problems}}  \\ 
         \cmidrule{3-18}
 
         & 
         & 
        \textbf{RL-F} & Sent.$\theta$	& \textbf{BS}	& CUI &
        \textbf{RL-F} & Sent.$\theta$	& \textbf{BS}	& CUI &
        \textbf{RL-F}& Sent.$\theta$	& \textbf{BS}	& CUI &
        \textbf{RL-F} & Sent.$\theta$	& \textbf{BS}	& CUI  \\ 
         \midrule  
         Rule-based & \textsc{Assmt} & \textbf{34.45} & 58.81	 & \textbf{59.80}	 & 38.97
                  & 12.31 &   \textbf{55.33}	 & 40.13	 &  34.23
                  & 9.49  &  \textbf{55.58}	 & \textbf{44.46} & 33.16
                  & 13.45 &  \textbf{68.61}	 & 50.32	 & 43.93
                  	  \\ 
                  	  \midrule 
                  	  
         \multirow{4}{*}{T5} & \textsc{Assmt} 
         & 32.77 & 59.57	 & 57.75 & 41.73
         & 13.68 & 53.44 & 39.72 &  \textbf{36.10}
         & 10.40 & 54.76	 & 44.16	 & 35.08
         & 14.82 & 67.49	 & 49.89	 & 44.51
           \\ 
         & \multicolumn{1}{r}{++}
        & 31.76 & 	58.74 & 57.12	 &  \textbf{42.19}
        & \textbf{13.78} & 	53.65 & \textbf{40.30}	 & 35.84
        & \textbf{10.55} & 54.10	 & 43.48	 & \textbf{35.20}
        & \textbf{15.00}  & 67.32	 & 50.36	 & \textbf{44.55}
            \\ 
          & \multicolumn{1}{r}{\textsc{A+Subj}}
         & 20.24 &  50.04   & 47.55     & 33.44
         & 9.52 & 51.91	 & 39.72	 & 30.43
         & 7.10 &  54.14	 & 43.87	 &  30.29
         & 10.89 & 64.63	 & 49.75	 & 39.02
         
          \\ 
           & \multicolumn{1}{r}{++}
         & 20.72   & \textbf{59.64}	 & 57.97     & 33.56
         & 9.46  & 53.55	 & 39.52	 & 18.76 
         & 7.35  & 54.69	 & 44.36	 & 14.40
         & 10.93  & 67.19	 & \textbf{50.42}	 &  24.83
           
           \\ 
           \midrule 
           \multirow{4}{*}{BART} & \textsc{Assmt} 
         & 25.70  & 54.98	 & 52.99	 & 32.49
         & 10.00  & 53.66	 & 39.08	 & 29.41
         & 8.04  & 54.66	 & 43.12	 & 29.04
         & 11.56  & 66.86	 & 48.48	 &  38.36
             \\ 
           & \multicolumn{1}{r}{++}
         & 28.22  & 57.04	 & 55.16	 &  32.28
         & 10.33  & 53.40	 & 39.21     &  30.75
         & 8.29    & 54.48	 & 44.01	 &  32.08
         & 11.65  & 66.67 	 & 49.23	 &  40.69
           \\ 
           & \textsc{A+Subj} 
         & 18.80  & 49.19	 & 46.77	 & 26.96
         & 7.04    & 51.70	 & 38.24	 & 25.30
         & 6.00    & 54.29	 & 43.71	 & 26.01
         & 9.25    &  64.95	 & 48.19	 & 34.02
             \\ 
            & \multicolumn{1}{r}{\textsc{++}} 
        & 20.23    & 57.91	 & 54.68	 & 32.91
        & 7.88    &  53.85	 & 40.21	 & 30.09
        & 6.85     & 54.61	 & 43.15	 & 30.12
        & 9.84    &  67.00	 & 49.70	 & 38.72
            \\ 
            \bottomrule 
    \end{tabular}
    
\end{adjustbox}
\vspace{-.1in}
    \caption{\small ROUGE-L F-score (RL-F), sentence embedding cosine similarity (\textit{Sent.$\theta$}), BERTScore (\textit{BS}), and evaluation using CUI F-score (\textit{CUI}) from fine-tuning T5 and BART on the two input settings: Assessment (\textsc{Assmt}), Assessment with Subjective sections(\textsc{A+Subj.}) ++ represents the training with data augmentation.   }
    \label{tab:result1semantic}
\end{table*}

\subsection{Evaluation}

We use ROUGE-L~\cite{lin2004rouge}, a conventional metric in summarization evaluation that based on n-gram overlap, as well as BERTScore~\cite{bert-score}, reporting maximum pairwise cosine similarity on word embedding from reference summary and predicted summary. ROUGE fails to recognize synonyms and abbreviations, which are common in biomedical text: e.g., \textit{heart attack} is the same clinical diagnosis as \textit{myocardial infarction}, and \textit{MI} is the abbreviation of \textit{myocardial infarction}. BERTScore compensates this limitation by using contextualized word embeddings from SapBERT~\cite{liu2021self}, a state-of-the-art BERT encoder~\cite{devlin2019bert} for biomedical entity representation that assigns high cosine similarity for synonyms and abbreviations based on UMLS. The reliability of both metrics are validated in literature, thus we report them as main results.  

Meanwhile, to better understand the system output, we provide two additional metrics that measure the quality of higher-level information and medical concepts. We took the hidden states of the last layer from SapBERT when taking reference and predicted summary as input, and measure the cosine similarity on sequence embedding (Sent.$\theta$). To evaluate the model's performance in predicting medical concepts, we ran QuickUMLS to get all CUIs from the reference and predicted summaries and computed the F-score. This metric has its own limitation due to the tricky parameter tuning in matching algorithms, causing superfluous or deficient extraction. Regardless, we include them as approximate solutions towards knowledge-based evaluation for clinical summarization, and leave the metric development for future work.  

In the experiments, we set the maximum input and output length to 512 and 128 tokens, respectively. The input text was truncated if the maximum length was exceeded. All experiments occurred on two NVIDIA Tesla V100 32GB GPUs. We used early stopping on the development set during training and saved the models with the highest validation ROUGE-L F-score for evaluation. More implementation details are presented in the Appendix~\ref{appendix:training}.


\begin{table*}[t]
\begin{adjustbox}{width=1\textwidth} 
\small 
    \centering
    \begin{tabular}{clllll|llll} \toprule
    
        \multirow{2}{*}{\textbf{Setting}} & \multirow{2}{*}{\textbf{Model}} & \multicolumn{4}{c|}{\textbf{Token Masking}} & \multicolumn{4}{c}{\textbf{Concept Masking}} \\ \cmidrule{3-10}
        & 
        & \textbf{RL-F} & Sent.$\theta$ & \textbf{BS} & CUI  
        & \textbf{RL-F} & Sent.$\theta$ & \textbf{BS} & CUI \\ 
        \midrule 
        \multirow{2}{*}{Explicit} & \textsc{Assmt}
        & 32.66 & 61.34$_{(\uparrow 2.53)}$ & 56.68  & \cellcolor{green!30}47.10$_{(\uparrow 8.13)}$
          & 29.86 & 55.87 & 53.91 & 40.27$_{(\uparrow 2.14)}$ \\ 
                 & \multicolumn{1}{r}{++} 
                 & 26.94 & 59.40$_{(\uparrow 0.59)}$ &  55.05 & 42.73$_{(\uparrow 3.76)}$
                 & 32.82 & 58.21 & 56.80 & 43.16$_{(\uparrow 4.19)}$ \\
                 \midrule 
                 
        \multirow{2}{*}{Direct} &\textsc{Assmt}
        & 12.69 &53.63 & 42.40$_{(\uparrow 2.27)}$ & 35.39$_{(\uparrow 1.16)}$
        & 14.90$_{(\uparrow 2.59)}$ & 55.48$_{(\uparrow 0.15)}$  & 47.10$_{(\uparrow 6.97)}$ & 35.29$_{(\uparrow 1.06)}$\\ 
        
          &\multicolumn{1}{r}{++} 
          & 10.44 & 53.47 & 43.46$_{(\uparrow 3.33)}$ & 37.45$_{(\uparrow 3.22)}$ 
         & \cellcolor{green!30}{15.76$_{(\uparrow 5.22)}$} & 56.82$_{(\uparrow 1.49)}$
         & \cellcolor{green!30}{48.72$_{(\uparrow 8.72)}$} & \cellcolor{green!30}37.74$_{(\uparrow 3.51)}$ \\ 
          \midrule 
          
           \multirow{2}{*}{Indirect}  & \textsc{Assmt} 
           & 10.07$_{(\uparrow 0.58)}$ & 52.72 & 41.47 & \cellcolor{green!30}38.19$_{(\uparrow 5.03)}$
          &  \cellcolor{green!30}{13.58$_{(\uparrow 4.36)}$} & 53.44 & 44.91$_{(\uparrow 0.45)}$  & 33.56$_{(\uparrow 0.40)}$ \\ 
           & \multicolumn{1}{r}{++} 	
           & 8.04 & 51.84 &40.45 & 37.53$_{(\uparrow 4.37)}$
           & 13.28$_{(\uparrow 4.06)}$ &55.02 & \cellcolor{green!30}{45.51$_{(\uparrow 1.05)}$} & 35.10$_{(\uparrow 1.94)}$ \\ 
           \midrule 
           
          \multirow{2}{*}{All} &  \textsc{Assmt} 	
          & 14.49$_{(\uparrow 1.04)}$ & 62.40  &49.62 & 40.44 
         & 18.72$_{(\uparrow 5.27)}$ &64.69 &54.03$_{(\uparrow 3.71)}$ & 42.69\\
         
           & \multicolumn{1}{r}{++} 
        & 12.12 & 63.08 & 50.20 & 
           \cellcolor{green!30}45.58$_{(\uparrow 1.65)}$
           & \cellcolor{green!30}{18.80$_{(\uparrow 5.35)}$} & 66.08 & \cellcolor{green!30}{55.29$_{(\uparrow 4.86)}$} & 44.56$_{(\uparrow 0.63)}$ \\ 
           
           \bottomrule
    \end{tabular}
     \end{adjustbox}
     \vspace{-.1in}
    \caption{\small Performance of T5 with domain adaptation pre-training using Assessment (\textsc{Assmt}) as input, under two mask policies: Token Masking and Concept Masking. We report Rouge-L F-score (RL-F), and BERTScore (BS), as well as Sentence embedding cosine similarity (Sent $\theta$) and CUI F-score. Numbers with\colorbox{green!30}{green background} address the highest performance across all results, with subscript number ($\uparrow$) denoting the improvements over rule-based results. }
    \label{tab:dapt_result}
   
\end{table*}

\section{Results and Analysis} \label{sec:results} 

We evaluated all systems on a test set of 92 progress notes and summaries. Recall that the progress notes contained the Subjective, Objective, Assessment and Plan sections. We set two types of input to the models: (1) Assessment section only (\textsc{Assmt}), (2) Assessment and Subjective sections (length permitting) (\textsc{A+Subj}). Both input settings also had augmented samples from the data augmentation method introduced earlier. We started with a simple rule-based system that was a UMLS concept extractor on the Assessment section. The evaluation metrics across the rule-based, fine-tuned T5 and BART (\cref{subsec:finetune}), and T5 with domain adaptation pre-training (DAPT, \cref{subsec:dapt}) are shown in Tables~\ref{tab:result1semantic} and \ref{tab:dapt_result}.  T5 with DAPT outperformed all other systems and established a benchmark performance for the task. We include a qualitative analysis to provide data-driven insights of the task (\cref{subsec:qualitative}). 

\subsection{Overall performance of fine-tuned models} \label{subsec:finetune} 
 
Table~\ref{tab:result1semantic} represents ROUGE-L F-score, cosine similarity on sentence embedding (\textit{Sent.}$\theta$), BERTScore and CUI F-score. Overall, scores dropped from Explicit Mentions to Direct Problems to Indirect Problems, likely due to increasing complexity with more abstractive concepts over extractive concepts. Explicit Mention summarization was the easiest and Indirect Problem summarization was the hardest. The rule-based system outperformed all T5 and BART variants on the Explicit Mentions, given that it identified the obvious entity mentions. 
For T5 and BART, fine-tuning with augmented samples slightly improved the ROUGE scores. Adding subjective sections (\textsc{A+Subj}) did not bring benefits, possibly because most subjective sections are empty in ICU progress note. 
T5 had more variants with better scores than BART. In our manual investigation, we find that BART generated text that are not relevant to medical domain\footnote{see Appendix~\ref{appendix:output} for example output for all models}. 
In sum, all fine-tuned model performance were tied with the baseline, which is impressive given that the baseline uses domain knowledge (medical concept). 

\begin{figure}[t]
    \centering
    \includegraphics[width=\columnwidth]{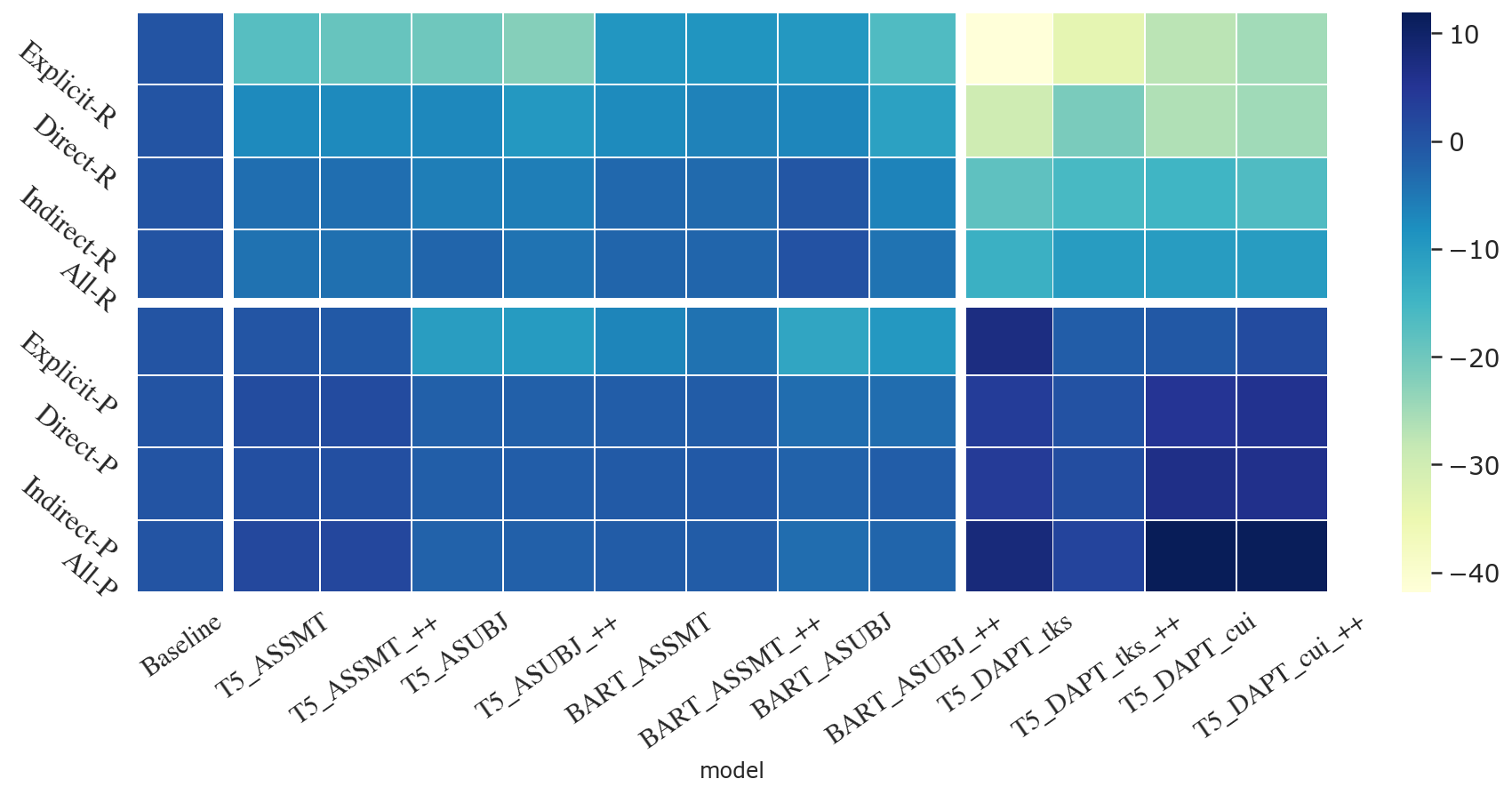}
    \vspace{-.33in}
    \caption{\small Performance drops (lighter color) and gains (darker color) over baseline (first column) on ROUGE-L Recall (top 4 rows) and Precision (bottom 4 rows). The darker the cell color is, the higher performance gain the model obtains over baseline. }  
    \label{fig:recall}
\end{figure}

\subsection{The effect of domain adaptation pre-training} \label{subsec:dapt}

Table~\ref{tab:dapt_result} contains results from training T5 with DAPT and fine-tuned on the annotated set across two methods for masking: random token (T5-DAPT-TKS) and concept masking (T5-DAPT-CUI). To highlight the differences before and after DAPT, we showed four scores as well as the performance gained over the baseline system on \textsc{Assmt} input. 
Overall, both DAPT settings delivered better performance. The performance gain of T5-DAPT-TKS was mainly from the CUI F-score (+1.16 to +8.13). Superior results were seen from T5-DAPT-CUI, achieving best performance on all setting except for Explicit Mention, yielding large performance gained on ROUGE F score(+2.59 to +5.35) and BERTScore (+0.45 to +8.72).  

In addition, Figure~\ref{fig:recall} includes the ROUGE Recall and Precision drops and gains from all models over baseline. ROUGE Recall measures the content coverage and Precision computes content relevancy in predicted summary~\cite{lin2004rouge}. All models reported lower recall compared to baseline, indicating their coverage was limited. T5 DAPT variants showed higher gains on precision, yielding the largest gain (+5 to +12 for T5-DAPT-CUI).   
These results indicate that continuously training T5 with domain vocabulary is a promising direction to solve the task. 

\begin{figure} 
\small 
    \centering
  \includegraphics[scale=0.52]{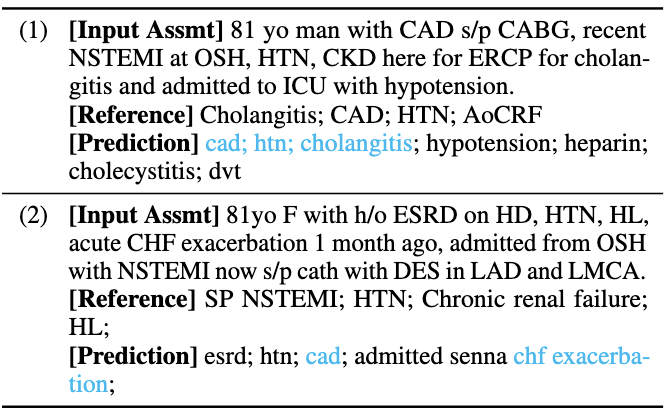}  
   \vspace{-.13in}
    \caption{\small Two cherry-picked examples from T5-DAPT-CUI output, with \textcolor{cyan}{cyan fonts} highlighted the correct diseases.\footnote{Semicolons are removed during fine-tuning and evaluation. We manually inserted them back for presentation purpose.}}
    \label{fig:chrrypick}
\end{figure}
\footnotetext[3]{Semicolons are removed during fine-tuning and evaluation. We manually inserted them back for presentation purpose.} 

\subsection{Qualitative analysis} \label{subsec:qualitative}
Besides the numeric metrics reported above, we provide example predicted summaries and qualitative analysis done by a domain expert (a critical care ICU physician). 
We cherry-picked two examples from T5-DAPT-CUI that best represent the characteristics of medical diagnostic consistency in clinical diagnostic reasoning, and present them in Figure~\ref{fig:chrrypick}. Example 6.1 shows the model performed extractive summarization: it generated both hypertension and hypotension as relevant diagnoses that represent an Indirect label for past medical history of hypertension, and Direct label for an active problem during the hospitalization with hypotension. In example 6.2, the model performed abstraction summarization. The last half of the Assessment highlights a type of heart attack (e.g., ``\textit{NSTEMI}") requiring an emergent medical procedure (e.g., ``\textit{cath wtih DES in LAD and LMCA}"), and the model summarized a rather complex statement into a single, accurate diagnosis of Coronary Artery Disease in its abbreviated form as ``\textit{CAD}".

\section{Discussion} \label{sec:discussion}



   
    


Our work begins with a single note in cross-sectional design to build our models; however, a patient's hospitalization is a multi-document workflow with repeated measures of progress notes and other note types across several days and multiple providers. In addition, providers generate their diagnoses via a reasoning process that includes structured data from vital signs, laboratory results, etc. Images and radiology reports are another modality that highlights the multi-modality approach in diagnostic reasoning. 
Nonetheless, our work opens the door for future research on knowledge intensive clinical summarization. This section includes a discussion of future directions in solving this task.  



\begin{figure}
\centering


\includegraphics[scale=0.55]{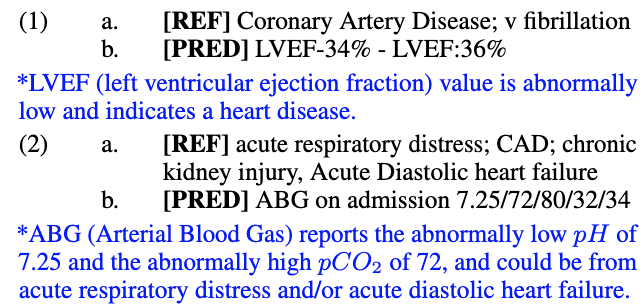}


\vspace{-.1in}
\label{fig:t5-all-2}
\caption{\small Two example reference (\textbf{REF}) and predicted summaries (\textbf{PRED}) from T5-ALL (input with objective sections).}
\end{figure}

\paragraph{Exploring structured data} The objective section of the progress note contain embedded structured data, delivering rich information regarding patient's problem. Recall the example in Figure~\ref{fig:task}, the reference summary contains diagnosis: ``\textit{Leukocytosis}" (high white blood cell count), ``\textit{anemia}"(low red blood cell count). These diagnosis are usually found in laboratory results. To investigate the use of objective sections and structured data, we append both Subjective and Objective sections in chronological order to the Assessment and input to T5 for fine-tuning and evaluation (T5-ALL), and we let the T5 tokenizer truncate the text when it exceeds the 512 token limit. On test set, the scores are too low to report. Yet, we observed that T5-ALL, instead of generating medical concepts, often extracts lines of lab values that strongly associate with the disease in reference summary (see Figure 7.1 and 7.2). This preliminary result indicates the future direction of understanding the association between disease and lab values in summarization.     

\paragraph{Incorporating knowledge into models}  We propose a knowledge intensive summarization task that requires clinical text understanding, knowledge representation and diagnostic reasoning. The experiment results showed that the models pre-trained on medical concepts effectively improved the performance, while challenges remain in understanding the associations among medications, symptoms and disease. Recent work on event extraction and clinical relation extraction incorporates biomedical knowledge graph into pre-trained language models~\cite{huang2020biomedical,roy2021incorporating}. Our future work will investigate the incorporation of knowledge graph into seq2seq pre-trained models.    

\paragraph{Evidence-based evaluation} Medical diagnosis is a critical component of effective healthcare but misdiagnosis is a major contributor to medical errors, especially in critical care settings where quick decision-making is needed. Medical diagnoses predicted by systems that are not redundant must be contextually relevant to the data gathered in a progress note to achieve valid reasoning. We believe an automated evaluation method for problem summarization should assess the knowledge representation, non-redundancy, and evidence relevancy, and the automated metrics used in our work cover partial aspects. Recently,~\citeauthor{moramarco2021towards} (\citeyear{moramarco2021towards}) studied a fact-based evaluation for medical summarization using human evaluation, which we plan to carry out in future work.

\section{Conclusion}
We propose a problem summarization task that address diagnostic reasoning, and show that T5 with DAPT achieves benchmarking performance for the task, but some key challenges remained. Our work lays the ground for future research on knowledge fused clinical summarizers as well as real-world clinical diagnostic decision support system. Future work will investigate the uses of structured data, evidence-based evaluation metric and better models for knowledge representation and summarization. 


\paragraph{Ethical Statement} The use of the data in this research came from a fully de-identified dataset (contains no protected health information) that we received permission for use under a PhysioNet Credentialed Health Data Use Agreement (v1.5.0). The study was determined to be exempt from human subjects research. 
All experiments followed the PhysioNet Credentialed Health Data License Agreement.

Medical charting by providers in the electronic health record is at-risk for multiple types of bias. Our research focused on building a system to overcome the cognitive biases in medical decision-making by providers. However, statistical and social biases need to be addressed before integrating our work into any clinical decision support system for clinical trials or healthcare delivery. In particular, implicit bias towards vulnerable populations and stigmatizing language in certain medical conditions like substance use disorders are genuine concerns that can transfer into language model training~\cite{thompson2021bias,saitz2021recommended,karnik2020structural}.
Therefore, it should be assumed that our corpus of notes for this task will carry social bias features that can affect fairness and equity during model training. Before the deployment of any pre-trained language model, it is the responsibility of the scientists and health system to audit the model for fairness and equity in its performance across disparate health groups~\cite{saleiro2018aequitas}. Fairness and equity audits alongside model explanations are needed to ensure an ethical model trustworthy to all stakeholders, especially patients and providers.

\bibliography{anthology,custom}
\bibliographystyle{acl_natbib}

\appendix

\section{Annotator Training}
\label{appendix:annotation}
We recruited two medical students as annotators who received training in medical school curriculum in SOAP note documentation. A three-week orientation and training was conducted by one of the critical care physicians. Each annotation achieved an inter-annotator agreement with a kappa score above 0.80 with the adjudicator. Another round of training was performed on 200 notes and the inter-annotator agreement was measured between annotators and the adjudicator. The annotation was reviewed if the kappa score is below 0.80 threshold.  

\begin{figure*}[ht!]
\small 
    \centering
  \begin{tabular}{p{\textwidth}}  \toprule 
  \textbf{Input}  
  ALCOHOL ABUSE   Assesment: This is a 37 year-old male with a history of alcoholism with  multiple admissions for detox and history of DTs/withdrawal seizures  who presents with intoxication followed by withdrawal. \\ 
  \textbf{Reference} ETOH intoxication; Elevated lipase \\
  \midrule 
   
  \textbf{FT-T5} ALCOHOL ABUSE Assesment: This is a 37 year-old male with multiple admissions for detox and history of DTs/withdrawal seizures who presents with intoxication followed by withdrawal.  \\ 
    \textbf{FT-BART} ALCOHOL ABUSE DIAGNOSIS   Assesment: This is a 37 year-old male with a history of alcoholism with DTs/withdrawal seizures. He presents with intoxication followed by withdrawal. Alcoholics Anonymous: He is an alcoholic with multiple admissions for alcohol abuse and \textcolor{red}{drug addiction. This male is the only person in the state of Alabama with alcohol dependence.}  \\ 
    \textbf{T5-DAPT-TKS} This is a 37 year-old male with history of DTs/withdrawal seizures who presents with intoxication followed by withdrawal \\ 
    \textbf{T5-DAPT-CUI} history alcoholism history history of dts/withdrawal seizures admissions alcohol abuse history \\
    
    \bottomrule

  \end{tabular}
  \caption{Given an input assessment, we show the reference summary, example output from fine-tuning T5 and BART, and T5 DAPT with token masking and concept masking. The red fonts show the information that is outside the input text.  }
  \label{fig:example_output}
  \end{figure*}

\section{Quality Measure for Data Augmentation}
\label{appendix:quality}

\begin{table}[]
\small 
    \centering
    \begin{tabular}{cccc} \toprule
        Input & Sent.$\theta$ & Jaccard &Length Diff.\\ \midrule 
       \textsc{Assmt} & 89.00 & 37.85 & 6.13 (4.12)  \\ 
       \textsc{Summ}  & 83.14 & 14.43 & 9.42 (5.99) \\ 
       \bottomrule 
    \end{tabular}
    \caption{\small Quality measurement on augmented input assessment (\textsc{Assmt}) and reference summary (\textsc{Summ}). For ever pair of original and augmented sample, we report cosine similarity between text embedding ($\theta$), Jaccard token overlap, and mean and standard deviation ($\sigma$) of length difference.}
    \label{tab:augmented}
\end{table}

\begin{table}[t]
\small 
    \centering
    \begin{tabular}{c|c} \toprule
      Hyper-parameter   & Setting   \\ \midrule 
      Optimizer   &  AdamW\\ 
      Epoch & 10 (with early stopping) \\ 
      Learning rate & 1e-3, 1e-4 \\ 
      Batch size & 256 \\ 
      Gradient accumulation & True \\ 
      \bottomrule
    \end{tabular}
    \caption{\small Hyperparameters for T5 DAPT}
    \label{tab:param_t5}
\end{table}

\begin{table}[t]
\small 
    \centering
    \begin{tabular}{c|c} \toprule
      Hyper-parameter   & Setting   \\ \midrule 
      Optimizer   &  Adam\\ 
      Epoch & 10 (with early stopping) \\ 
      Learning rate & 1e-4, 1e-5, 1e-6 \\ 
      Batch size & 4 \\ 
     Task Prefix (t5) & ``summarize:" \\ 
     Encoder max length & 512 \\ 
     Decoder max length & 128 \\ 
     Beam size & 10 \\ 
     Length penalty & 1 \\ 
     no repeat ngram size & 2 \\ 
      \bottomrule
    \end{tabular}
    \caption{\small Hyperparameters for fine-tuning T5 and BART}
    \label{tab:param_ft}
\end{table}

The quality of augmented data directly affected the training process. To ensure a high-quality training corpus, we randomly selected 2,000 pairs of augmented samples. We evaluated how well the meanings were preserved in the augmented sample, and how much lexical variance was introduced into the augmented samples. We reported cosine similarity between the embedding pairs for quality of meanings, and we reported Jaccard similarity for degree of string overlap. Specifically, given a pair of the original sample and the augmented sample, we generated a text embedding through SapBERT~\cite{liu2021self}, a BERT encoder pre-trained to represent biomedical entities using UMLS. We expected a high cosine similarity if the augmented samples expressed the same meanings as the original samples. We ran Jaccard similarity by treating the samples as lists of tokens, and expected a low Jaccard score if there were new terms introduced in the augmented samples, e.g. \textit{ARF} and \textit{Acute Renal Failure}. We also reported the mean and standard deviation of the length differences between original and augmented samples (Table~\ref{tab:augmented}). On both input assessment and reference summary, the cosine similarity between original and augmented samples was higher than 0.80. Assessment input contained more words that were not biomedical concepts; thus, the augmented sample had a greater proportion of overlapping text than the reference summary. Both had more than 6 token differences in length. In conclusion, our proposed strategy of data augmentation successfully produced a high quality training corpus.

\section{Hyperparameters}
\label{appendix:training}


Here we report the hyper-parameters we used for T5 DAPT experiments in table~\ref{tab:param_t5}, and fine-tuning for t5 and BART in table~\ref{tab:param_ft}. The input length to both T5 and BART is set to 512 tokens. On the training data, the average length of input assessment is 43.33 tokens, and the average length of input and subjective sections is 70.97 tokens. Therefore the maximum encoder length is appropriate for our task.

\section{Model Example Output}
\label{appendix:output}
Figure~\ref{fig:example_output} shows the example output from fine-tuning T5 and BART, and T5-DAPT with token masking as well as concept masking policy. T5-DAPT-CUI extracts medical concepts. FT-T5 and T5-DAPT-Tks extract sequence of text from input assessment. FT-BART produces text with information that is not mentioned in the input (red fonts).    

\label{sec:appendix}

\end{document}